\documentclass[DTMColor]{ROB-New}
\usepackage{amsmath,amssymb,amsfonts}
\usepackage{algorithmic}
\usepackage{graphicx}
\graphicspath{{Illustration_eps/}}
\usepackage{textcomp}
\usepackage{soul, color, xcolor}
\usepackage{makecell}
\usepackage{array} 
\usepackage{stfloats}
\usepackage{geometry}
\usepackage{comment}
\usepackage{multirow}
\usepackage{booktabs}
\usepackage{mwe}
\usepackage{afterpage}

\usepackage{colortbl}

\soulregister{\cite}7 
\soulregister{\citep}7 
\soulregister{\citet}7
\soulregister{\ref}7 
\soulregister{\pageref}7 

\newcommand{\orcids}[1]{\href{https://orcid.org/#1}{\includegraphics[width=10pt]{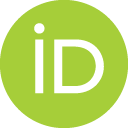}}}

\begin{document}

\authormark{Ningquan Gu et al.}

\articletype{RESEARCH ARTICLE}

%\jmonth{January}
\jnlPage{1}{17}
% \jyear{2024}
\jdoi{10.1017/S0263574723001753}

\title{ShakingBot: Dynamic Manipulation for Bagging}

\author[1]{Ningquan Gu} \orcid{0000-0002-7032-900X}
\author[1]{Zhizhong Zhang}
\author[1]{Ruhan He\hyperlink{corr}{*}}
\address[1]{School of Computer Science and Artificial Intelligence, Wuhan Textile University}

\author[2]{Lianqing Yu}
\address[2]{School of Mechanical Engineering and Automation, Wuhan Textile University}

\address{\hypertarget{corr}{*}Corresponding author. \email{heruhan@wtu.edu.cn}}

\received{09 March 2023}
\revised{09 November 2023}
\accepted{03 December 2023}


\abstract{
	Bag manipulation through robots is complex and challenging due to the deformability of the bag.
	Based on the dynamic manipulation strategy, we propose a new framework, ShakingBot, for the bagging tasks.
	ShakingBot utilizes a perception module to identify the key region of the plastic bag from arbitrary initial configurations. 
	According to the segmentation, ShakingBot iteratively executes a novel set of actions, including Bag Adjustment, Dual-arm Shaking, and One-arm Holding, to open the bag. 
	The dynamic action, Dual-arm Shaking, can effectively open the bag without the need to take into account the crumpled configuration.
	Then, the robot inserts the items and lifts the bag for transport. 
	We perform our method on a dual-arm robot and achieve a success rate of 21/33 for inserting at least one item across various initial bag configurations.
	In this work, we demonstrate the performance of dynamic shaking action compared to the quasi-static manipulation in the bagging task.
	We also show that our method generalizes to variations despite the bag's size, pattern, and color.
	Supplementary material is available at \href{https://github.com/zhangxiaozhier/ShakingBot}{https://github.com/zhangxiaozhier/ShakingBot}.}

\maketitle

\section{Introduction}
Plastic bags are ubiquitous in everyday life, including in supermarkets, homes, offices, and restaurants. 
Bagging is a helpful skill, including opening a bag and inserting objects for efficient transport.
Therefore, handling bagging tasks is meaningful and practical.
However, the task is challenging for the robot because of the inherent complexity of thin plastic dynamics and self-occlusions in crumpled configurations.

Prior works concerning deformable object manipulation are considerable. 
Most of the research focuses on manipulating linear objects \cite{zhang2021robots,she2021cable,wang2019learning,lim2021planar,zhu2019robotic} and fabric \cite{weng2022fabricflownet,mo2022foldsformer,chen2022efficiently,hoque2020visuospatial,lin2022learning}.
% The manipulation of more deformable objects, such as bags and sacks, has received limited attention because of its increased complexities. 
% Among the relevant works in this area, Chen et al. \cite{chen2022autobag} proposed the AutoBag algorithm to manipulate a robot to open bags, insert items, lift them, and transport them to a target zone. 
Among the most recent manipulation of more deformable objects, such as bags and sacks, Chen et al. \cite{chen2022autobag} proposed the AutoBag algorithm to manipulate a robot to open bags, insert items, lift them, and transport them to a target zone. 
% Their approach employed quasi-static interactions to expand the bag and insert items. 
However, their quasi-static action method required a significant number of interactions.
%, particularly when dealing with challenging initial configurations such as highly crumpled bags.

Dynamic manipulation \cite{mason1993dynamic} is a common and efficient action in real life, which can dramatically reduce the complex configuration of the object without taking into account the initial complex state. For example, in our daily lives, people will open the crumpled bag by grasping the bag's handles and shaking their arms up and down to get air into the plastic bag without considering the crumpled configurations of the bags.
Then, we consider applying the dynamic manipulation strategy to the bagging task with a dual-arm robot.

Reinforcement Learning and Learning from Demonstration have demonstrated success in cloth manipulation tasks \cite{hietala2022learning, wu2019learning, jangir2020dynamic, jia2019cloth,lee2021learning}. 
% For instance, Seita et al. \cite{seita2020deep} employed deep imitation learning to achieve sequential fabric smoothing, while Canberk et al. \cite{canberk2022clothfunnels} utilized reinforcement learning to unfold and smooth cloth.
However, they often require substantial data and are frequently trained in simulation.
% When manipulating plastic bags, building dynamics models becomes challenging due to the complex non-linear dynamics involved. 
% The intricate interactions between aerodynamics and bag deformations further compound the difficulty of collecting accurate datasets, regardless of whether they are obtained from simulations or real-world environments.
It's more challenging for plastic bags to build dynamic models and collect accurate datasets whether from simulation or real-world environments.
Therefore, we utilize the idea of most visual policy learning approaches \cite{gao2023iterative,chen2022autobag} to guide the manipulation with action primitives.

\begin{figure*}[t]
	\centerline
	{
		\includegraphics[
		width=0.8\textwidth]
		{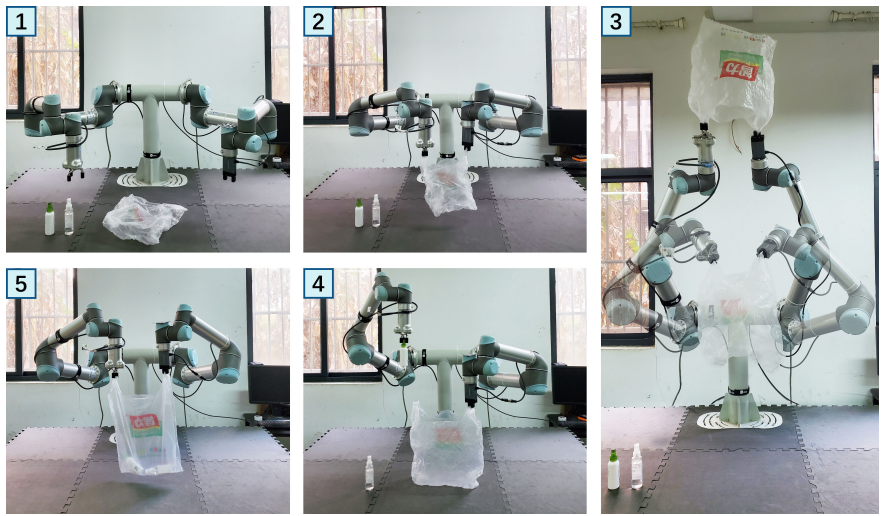}
	}
	\caption{\textbf{Opening the bag with dynamic actions.} 
		(1) Initial highly unstructured bag and two solid objects. 
		(2) Through region perception, the robot grasps the two handles and adjusts the distance between the two arms.
		(3) The arms shake the bag at high speed according to the pre-defined trajectory, which makes the air into the bag. 
		(4) One arm holds the opened bag on the workspace. 
		(5) The two arms lift the bag filled with the inserted items
	}
	
	\label{fig-shakingbot}
\end{figure*}

In this work, we focus on bagging tasks, including opening the bag and inserting items, where the key goal is to maximize the opening area of the bag. An ideal bagging approach should be:
\begin{itemize}

	\item Efficient: The approach should reach enough opening area with a few actions from arbitrarily crumpled initial configurations to insert items.
	
	\item Generalizable: The algorithm should generalize to different colors, patterns, and sizes.
\end{itemize}

To achieve this goal, we present a new framework, ShakingBot, for manipulating a plastic bag from an unstructured initial state so that the robot can recognize the bag, open it, insert solid items into it, and lift the bag for transport.
Given an initial highly unstructured bag and two solid objects, we train a perception module to recognize the key regions of the bag, where we reduce the workload through the colored bags to get the training dataset. 
After grasping the recognized handles, the robot iteratively executes a novel set of actions, including Bag Adjustment, Dual-arm Shaking, and One-arm Holding, to open the bag. 
Through dynamic action, we can be very effective in opening the bag.
When the bag opening metric exceeds a threshold, ShakingBot proceeds to the inserting item stage.
The simplicity of ShakingBot, combined with its superior performance over quasi-static baselines, emphasizes the effectiveness of dynamic manipulation for bagging tasks.
See Figure~\ref{fig-shakingbot} for ShakingBot in action on a dual-arm robotic (consisting of two UR5 robotic arms).

In summary, the main contributions of this paper are:

\begin{enumerate}
	\item  Based on the dynamic manipulation strategy, we propose a new framework, ShakingBot, for the bagging tasks, which improves the success rate and efficiency compared to the quasi-static strategy.
	\item We design a novel set of action primitives for dynamic manipulation, including Bag Adjustment, Dual-arm Shaking, and One-arm Holding, which make it possible to apply dynamic manipulation in bagging tasks.
	\item Experiments to demonstrate that ShakingBot generalizes to other bags with different sizes, patterns, and colors, even without training on such variations.
\end{enumerate}

\begin{figure*}[!t]
	\centerline
	{
		\includegraphics[
		width=1\textwidth
		]{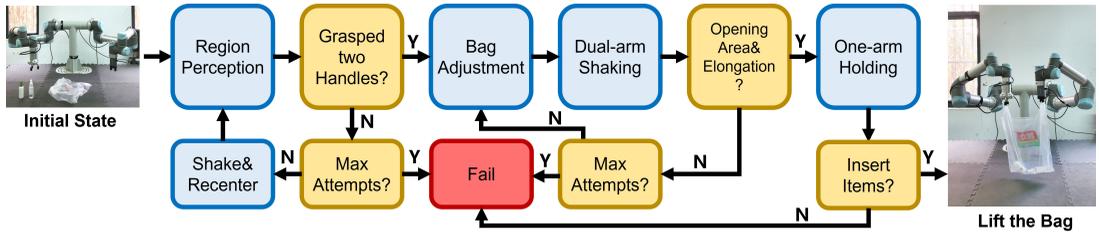}
	}
	
	\caption{ \textbf{Overview of ShakingBot.}
		See the left of the figure. The robot starts with an unstructured bag and two items.
		As shown in the flow, ShakingBot opens the bag according to the steps shown (see Section \ref{shakingbot_bagging} for details).
		When the bag opening metric exceeds a certain threshold, ShakingBot proceeds to the item insertion stage.
		If the robot lifts the bag with all the items inside, the trial is a complete success}
	\label{fig-overview}
\end{figure*}

\section{Related work}
\subsection{Deformable object manipulation}
Deformable object manipulation remains challenging for robots, primarily due to the complex dynamics and the limitless set of possible configurations.
There have been significant prior works in the manipulation of controlling ropes \cite{nair2017combining,wang2019learning,zhang2021robots}, cables \cite{lim2021planar,nakagaki1996study,nakagaki1997study,she2021cable,zhu2019robotic}, smoothing fabric \cite{chen2022efficiently,ha2022flingbot,hoque2020visuospatial,lin2022learning} and folding cloth \cite{weng2022fabricflownet,hietala2022learning,gu2023defnet,mo2022foldsformer}.
% However, the manipulation of more deformable objects, such as bags and sacks, has been limited due to the increased complexities associated with deformable objects.
However, these methods cannot be generalized to increased complexities of deformable objects, such as bags and sacks.
For example, Weng et al. \cite{weng2022fabricflownet} employed the optimal-flow method to predict the movement of each fabric particle to calculate pick-and-place points. Yet, their method can only handle one-step manipulation and requires knowledge of the object's next state. It is difficult to know the comprehensive information regarding the state of the bag in each step and to predict the movement of each bag particle.
Mo et al. \cite{mo2022foldsformer} proposed a sequential multi-step approach with space-time attention for cloth manipulation. However, their approach focuses on predicting intermediate states during multi-step manipulation with limited configuration, which cannot be applied to arbitrary configurations of a bag.
On the other hand, the early research work of bag-shaped objects focused on the mechanical design of robots capable of grasping \cite{kazerooni2005robotic} or unloading \cite{kirchheim2008automatic} sacks.
Some studies paid attention to zip bags \cite{hellman2018functional} in constrained setups or supposed that a sturdy, brown bag was already open for item insertion, as with a grocery checkout robot \cite{klingbeil2011grasping}.
In recent work, Gao et al. \cite{gao2023iterative} proposed an algorithm for tying the handles of deformable plastic bags.
While their approach modeled bags with a set of 19 manually labeled key points, it is challenging to estimate the representation with a bag in highly unstructured configurations.
Chen et al. \cite{chen2022autobag} proposed an AutoBag algorithm to open the bag and insert items into the bags, where the bags began empty and in unstructured states. They defined a novel set of quasi-static action primitives for manipulating bags.
However, their approach was not validated on bags with different colors and required many actions to open the bag. In contrast, we focus on the bagging tasks, which can generalize to different colors, and we are able to execute fewer actions to finish the task.

\subsection{Dynamic manipulation}
As opposed to quasi-static manipulation, dynamic manipulation \cite{mason1993dynamic} uses the acceleration forces produced by robots to manipulate objects.
By building up momentum with high-velocity actions, the system can manipulate out-of-contact regions of the deformable object.
A great deal of progress has been made in the field of deformable dynamic manipulation.
Examples include high-speed cable knotting \cite{hopcroft1991case,morita2003knot}, cable manipulation with fixed endpoints using learning-based approaches \cite{zhang2021robots} or a free endpoint \cite{zimmermann2021dynamic,chi2022iterative}.
Recently, Ha and Song proposed the FlingBot, which learned to perform dynamic fling motions with a fixed parameter to smooth garments using Dual-UR5 arms. Their approach was firstly trained in SoftGym simulation \cite{lin2021softgym} and fine-tuned in the real world to get the learned grasp points with two UR5 robots.
The results suggested significant efficiency benefits compared to quasi-static pick-and-place actions. 
However, their research object is fabric, and their action primitives are not suitable for bagging tasks.
In terms of bag manipulation, Xu et al. \cite{xu2022dextairity} produced high-speed interactions using emitted air in order to open the bag. However, they required more equipment. They used a setup with three UR5 robots to control two grippers and a leaf blower.
In contrast, we apply dynamic manipulation to the bagging tasks with less equipment. By executing a novel set of dynamic action primitives with two UR5 robots, we can achieve high performance from crumpled initial bag configurations.

\subsection{Learning for deformable object manipulation}
Concerning learning methods for deformable object manipulation, Learning from Demonstration (LfD) and Reinforcement Learning (RL) have demonstrated success in cloth manipulation tasks \cite{hietala2022learning, wu2019learning, jangir2020dynamic, jia2019cloth}. For instance, Seita et al. \cite{seita2020deep} employed deep imitation learning to achieve sequential fabric smoothing, while Canberk et al. \cite{canberk2022clothfunnels} utilized reinforcement learning to unfold and smooth cloth.
However, it is important to note that these methods often require a substantial amount of data, which can be challenging to collect in real-world environments due to factors such as wear and tear on the robot and limitations in available observation methods. As a result, they are frequently trained in simulators \cite{brockman2016openai,lin2021softgym,todorov2012mujoco}.
When it comes to manipulating plastic bags, building dynamics models is challenging due to the complex non-linear dynamics involved. Besides, obtaining accurate datasets for plastic bag manipulation proves particularly difficult, regardless of whether they are obtained from simulations or real-world environments. This challenge is exacerbated by the thinness of the bags and their interactions with the surrounding air.
Considering these challenges, we chose a more direct approach for our bagging task \cite{gao2023iterative,chen2022autobag}. 
Leveraging visual policy learning approaches to detect the key points of the bag and adopt the method of action primitives to tackle the task. 
In order to incorporate dynamic manipulation into the bagging task, we design action primitives to overcome the potential challenges that may arise during dual-arm manipulation. Moreover, we conduct ablation experiments to demonstrate the effectiveness of these primitives.

\section{Problem statement}
The bagging task is formulated as follows: First, a bag is placed on a flat surface with random configurations. Second, we need to open the bag. Lastly, we insert $n$ items and lift the bag for transport.

\begin{figure}[htbp]
	\centerline
	{
		\includegraphics[
		width=0.6\textwidth]{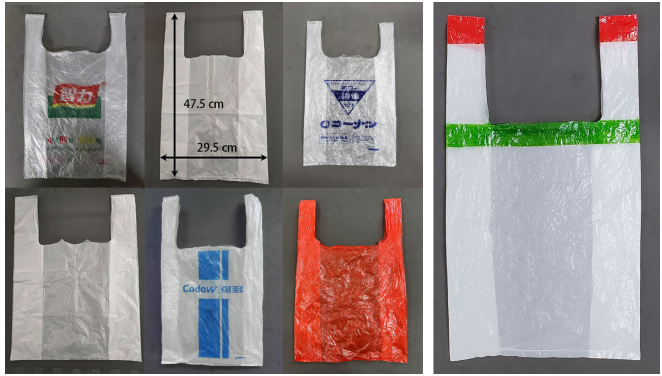}
	}
	\caption{Left: Various plastic bags adopted to train and test the region perception module. The bags include different sizes, patterns, and different colors. Right: A bag with red paint on its handles and green paint around its rim. The paint color can be changed into others according to the pattern color of the bag
	}
	\label{fig-bags}
\end{figure}

We use the most commonly used plastic bag, the vest plastic bag, as our research object. We define two labels for a bag, ``handle'' and ``rim'' (see the right of Figure~\ref{fig-bags}). The ``handle'' is the point where to grasp by the robot.
The rim surrounds the opening area, and its opening orientation is determined by the direction of the outward-pointing normal vector from the plane formed by the opening. 
The larger the opening area, the easier it is to put items in. In random configurations, the direction and area of the opening are various.
It is assumed that the bag's initial state is unstructured: deformed, potentially compressed. 
The handles and rim of the bag may be in a partial or fully occluded configuration.
We need to manipulate the bags with different colors, patterns, and sizes. 
See the left of Figure~\ref{fig-bags} for an illustration.
We do not consider the case of a brand-new bag without any deformation and wrinkles, the configurations of a brand-new bag are almost known, and the two sides of the bag stick to each other tightly without any space. 
This is an extremely special case.

We consider a bimanual robot (consisting of two UR5 robotic arms) with grippers and a calibrated camera above a flat surface. 
Let $\pi_{\theta}:\mathbb{R}^{W\times H} \rightarrow \mathbb{R}^{2}$ be a function parameterized by $\theta$ that maps a depth image $\mathbf{I}_{t} \in \mathbb{R}^{W\times H }$ at time $t$ to an action $\mathbf{a}_{t}=\pi_{\theta}\left(\mathbf{I}_{t}\right)$ for the robot. 
In order to simplify the question, we constrain the experimental area $\mathbb{R}^{2}$ to be within reach of the robot.
We assume a set of rigid objects $\mathcal{O}$, placed in known poses for grasping. 
We are interested in learning the policy $\pi$ such that the robot can open the bag. 
After opening the bag, we insert the objects into the bag. Lastly, we lift the bag off the table while containing the objects for delivery.

\section{Approach}
\subsection{Method overview}
Bagging aims to open the bag from an arbitrarily crumpled initial state and insert the object.
Concretely, this amounts to maximizing the opening area of the bag on the workspace to make it easier to put objects in. 
It is intuitive that dynamic actions can appropriately make use of the airflow through a high-velocity action to open the bag, and then it can achieve high performance on bagging tasks.

\begin{figure*}[htbp]
	\centerline
	{
		\includegraphics[
		width=0.8\textwidth]{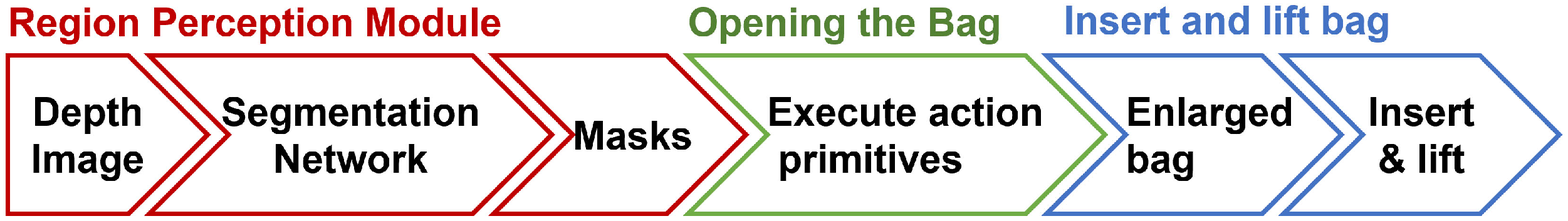}
	}
	\caption{\textbf{Pipeline for our method:} The perception module takes depth images and outputs segmentation masks for the bag handles and rim. The robot grasps the key points and executes dynamic actions. Last, the robot inserts the items and lifts the bag
	}
	\label{fig-pipeline}
\end{figure*}

Figure~\ref{fig-pipeline} provides the overall pipeline of our method.
We propose a learned region perception module to recognize the bag handles and the rim (Section \ref{segmentation}). 
% The training dataset is collected through the color-labeling process (Section \ref{segmentation}).
We define a novel set of action primitives (Section \ref{action}), including Bag Adjustment, Dual-arm Shaking, and One-arm Holding, for dynamic manipulation. 
We then describe the ShakingBot framework (visualized in Figure~\ref{fig-overview}) to open the plastic bag and insert objects (Section \ref{shakingbot_bagging}).

\subsection{Region perception module} \label{segmentation}
In this module, we utilize semantic segmentation to identify important regions of the bag, including the grasp points and the opening area.
We define the handles of the bag as the grasping points, while the rim of the bag is used to calculate the opening area.
The neural network is trained by the depth images of the scene containing the bag.
By predicting semantic labels for each pixel point, the network gives the probability that the pixel contains the handles and rim of a bag.
Lastly, we can obtain semantic segmentation masks of the bag.

In this paper, we employ DeeplabV3+ \cite{chen2018encoder} as the segmentation algorithm for our region perception module. 
The rationales behind this choice are demonstrated in Section~\ref{network-compare}.

\begin{figure}[htbp]
	\centerline
	{
		\includegraphics[
		width=0.8\textwidth]{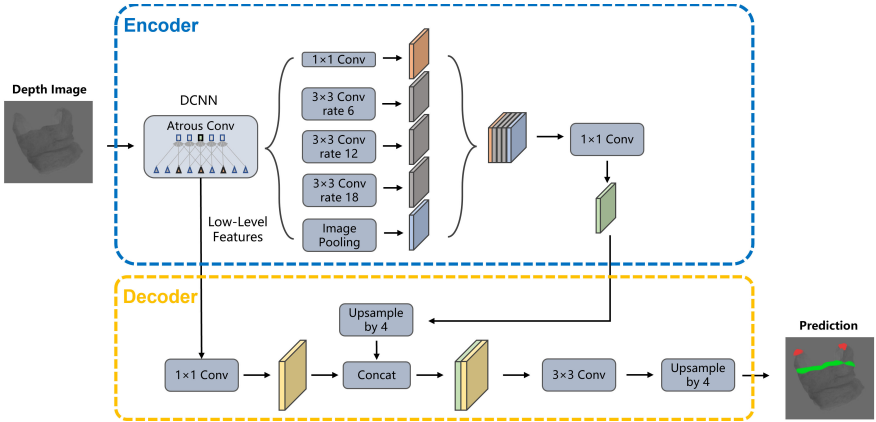}
	}
	\caption{\textbf{The architecture of Deeplabv3+}
	}
	\label{fig-deeplabv3plus}
\end{figure}

DeepLabV3+  \cite{chen2018encoder}  is an advanced semantic image segmentation model and the latest version in the DeepLab series.
It consists of two parts: the Encoder and the Decoder, see Figure~\ref{fig-deeplabv3plus}.
The Encoder part consists of a DCNN backbone network and an Atrous Spatial Pyramid Pooling (ASPP) module in a serial structure. The DCNN backbone, which utilizes the ResNet model, is responsible for extracting image features. The ASPP module processes the output of the backbone network using a 1x1 convolution, three 3x3 dilated convolutions, and a global pooling operation. The input to the Encoder is the depth image, and it produces two outputs. One output is directly passed to the Decoder, while the other output is processed through the ASPP module and then concatenated. A 1x1 convolution is applied to compress the features and reduce the number of channels.
The Decoder part takes the intermediate output of the DCNN backbone and the output of the ASPP module and transforms them into the same shape. These outputs are then concatenated and further processed through a 3x3 convolution and upsampling operations. The final result is obtained after these operations.

In our region Perception module, we define the loss $\mathcal{L}(o, t)$ to be the mean of the point-wise binary cross-entropy loss for every class $k$, $k \in K$:

\begin{equation}
		\mathcal{L}(o, t)=  \frac{1}{K} \sum_k^K (-\frac{1}{N} \sum_{i} w_i *(t[i] * \log (o[i])  +(1-t[i]) * \log (1-o[i])))
\end{equation}

Where $o[i]$ is the prediction segmentation at the image location $i$ for the class $K$, while the $t[i]$ is the ground truth.
We add a weight $w$ to deal with the problem of imbalanced distribution between the positive and negative labels.

As we showed in Section \ref{shakingbot_bagging}, the perception allows us to grasp the handle and calculate the opening area.

\subsection{Action primitive definition} \label{action}
In order to achieve an efficient, generalizable, and flexible bag-opening system, we propose that the system contains two arms that can operate in a dynamic action space. 
% We also consider the process of dual-arm bag manipulation, it's vital to overcome the potential challenges, for example, getting more air into the bag, preventing the bag from sticking to each other, making full use of the movement space for the dual-arm robot, and prevent the opened bag from collapsing during inserting items.
% To this end, we design a novel set of action primitives, including Bag Adjustment, Dual-arm Shaking, and One-arm holding.
%, for a dual-arm robotic, where the two arms are placed on either side of the workspace to avoid interacting with each other.
Additionally, we address the potential challenges associated with dual-arm bagging manipulation, such as ensuring sufficient airflow into the bag, preventing bags from sticking together, maximizing the dual-arm robot's range of movement, and preventing the opened bag from collapsing during item insertion.
To overcome these challenges, we introduce a novel set of action primitives, including Bag Adjustment, Dual-arm Shaking, and One-arm Holding. These novelly defined action primitives enable dynamic manipulation in bagging tasks.
% These defined novel action primitives make it able to apply dynamic manipulation to bagging tasks.
The primitives are as follows.

\subsubsection{\textbf{Bag Adjustment} $(d, \Delta d, k_s, l, f)$} \label{bag_adjustment}

\begin{figure}[htbp]
	\centerline
	{
		\includegraphics[
		width=0.6\textwidth]{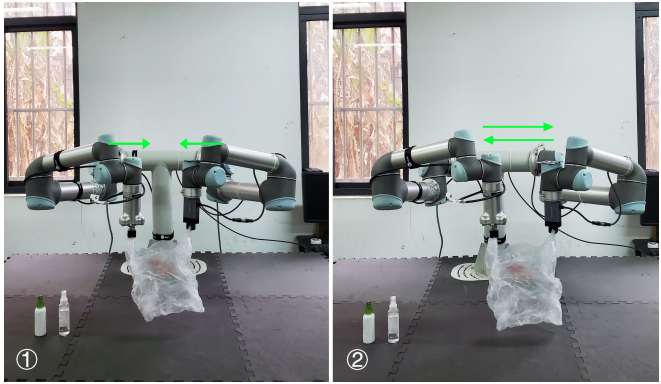}
	}
	\caption{ \textbf{Bag Adjustment:} \textcircled{1}The two arms decrease their distance. \textcircled{2}Robot swings the bag in the horizontal left and right directions
	}
	\label{fig-adjustments}
\end{figure}

After grasping the handles, the two grippers maintain the position left and right symmetrical.
We decrease the initial distance $d$ between the two grippers by $\Delta d$.
This can increase the chances of getting more air into the bag during the dynamic shaking action, which is beneficial to enlarge the bag.
But, if the distance between the arms is less than the threshold, the action doesn't execute.
After the action, the robot performs $k_s$ times swinging movements with length $l$ and frequency $f$ in the horizontal right and left directions. 
It can separate the two layers of the rim, which is beneficial for preventing them from sticking to each other.
(See Figure~\ref{fig-adjustments})

The Bag Adjustment primitive could increase the success rate of subsequent Dual-arm Shaking attempts to enlarge the bag more efficiently.

\subsubsection{\textbf{Dual-arm Shaking} $(H,H^{'},v)$}
\begin{figure}[htbp]
	\centerline
	{
		\includegraphics[
		width=0.6\textwidth]{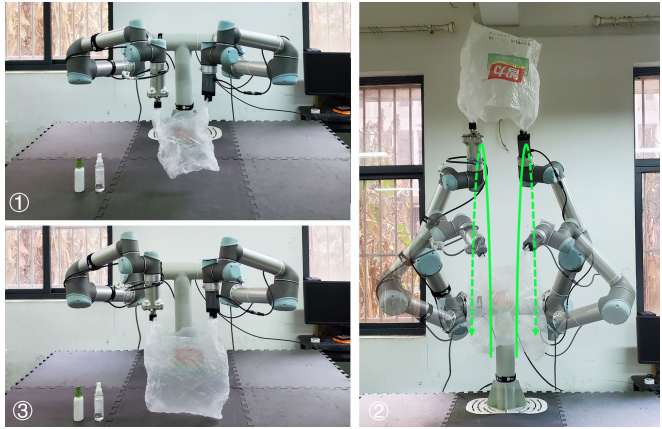}
	}
	\caption{\textbf{Dual-arm Shaking:} \textcircled{1} Preparation for dynamic dual-arm shaking. \textcircled{2}Long-distance and high-speed shaking. \textcircled{3} An enlarged bag
	}
	\label{fig-dynamic-shaking}
\end{figure}

We define the dynamic shaking action as follows:
The dual arms shake the bag dramatically from the current position to the height of $H$ at $v$ velocity in the vertical direction, then pull down to the height of $H^{'}$ with the same velocity.
The direction of the grippers should change from the downward direction at the initial position to the $45^{\circ}$ direction from the horizontal at the highest position. When the grippers return to the height of  $H^{'}$, the direction of the grippers returns vertically downward. It can make better use of the bag's acceleration and movement space by changing the direction of the grippers.
We set the $H$=1.4m to give the bag plenty of movement distance. It is beneficial for making more air into the bag to enlarge the opening. 
The $H^{'}$ should be higher than the bag's bottom to prevent the bag from touching the surface. The RGBD camera can detect the height of the bag's bottom.
The setting of primitive could make full use of the movement space to open the bag during shaking.
(See Figure~\ref{fig-dynamic-shaking})

\subsubsection{\textbf{One-arm Holding} $(h)$} \label{one_arm_holding}
\begin{figure}[htbp]
	\centerline
	{
		\includegraphics[
		width=0.6\textwidth]{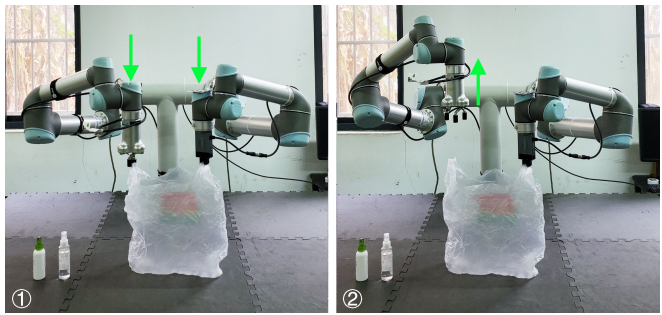}
	}
	\caption{\textbf{One-arm Holding: }\textcircled{1}Move the bag to touch the surface of workspace. \textcircled{2} The gripper releases the right handle while the bag keeps its shape
	}
	\label{fig-place}
\end{figure}

The dual arms, which grasp the bag, move down to the height of $h$, and the bottom of the bag just touches the workspace. 
The $h$ is determined by the height of the bag's bottom.
After that, the right gripper releases the handle and moves away.
The workspace surface and the left gripper support the bag together, which can prevent the bag from deforming itself when inserting items.
(See Figure~\ref{fig-place})

In addition to the above action primitives, we also adopt some primitives from \cite{chen2022autobag}, including Shake and Recenter primitives.
The difference between our Dual-arm Shaking and their Shake is that our method is a dual-arm action with a significant movement distance in the vertical direction to get air into the bag. 
In contrast, their action is to rotate one of the robot's wrists side by side to expand the bag's surface area.

\subsection{ShakingBot: bagging task} \label{shakingbot_bagging}
Firstly, the ShakingBot utilizes the perception module to recognize the positions of the handles to grasp them. Secondly, it chooses actions to execute according to the bag opening metrics \cite{chen2022autobag}. 
Finally, the ShakingBot inserts the items into the bag and lifts it. 
See Figure~\ref{fig-overview} for an overview.
The ShakingBot consists of the following three steps:

\subsubsection{Grasping the handles}
We input the depth image to the region perception module and get the predicted handle segmentation. The center of each handle is the grasp point. 
If the two handles region of the bag can be recognized, the robot grasps the two key points directly.
Otherwise, the robot executes a Shake. 
The Shake action can expand the area of the bag, and its grasp point is obtained from the handle region (if handles are not visible, we select anywhere on the bag as the grasp point).
If the bag is not in the center of the camera frame, we execute the Recenter action. 
After these actions, we execute region perception and grasping action again. If the two handles still can not be grasped, the robot repeatedly executes the above actions.
The grasping height is set to the height of the workspace in order to ensure that the handles can be grasped successfully.
% because the depth values are unreliable due to the bag's reflective material. 
The grasp position coordinates are specified as Cartesian coordinates in pixels. 

\subsubsection{Opening the bag}
After grasping the two handles of the bag, the two grippers move the bag to a fixed height with a pre-set distance between the two grippers.
And the bag is still in a crumpled configuration.
The robot applies Bag Adjustment and Dual-arm Shaking primitives to enlarge the bag iteratively.
During each iteration, our algorithm observes an overhead image of the bag to estimate its rim position and adopts the two opening metrics, Normalized convex hull area $A_{CH}$ and Convex hull elongation $E_{CH}$ \cite{chen2022autobag}, to evaluate the enlarged results. 
Repeating this process until the normalized convex hull area reaches a threshold $A_{CH}$ value and the elongation metric falls below a threshold  $E_{CH}$ value, which means that the opening is large enough for inserting objects.

\subsubsection{Inserting and lifting}
The robot performs the One-arm Holding action to place the bag and estimates the openings by fitting convex hulls on rims. We divide the opening spaces by the number of objects.
Later, the gripper grasps the objects that are placed in known poses, and the robot places them in the center of each divided region. 
After performing these actions, the robot identifies the position of the released handle and executes the grasping action. 
In the end, the robot lifts the bag from the table.

\section{Experiments}
In training and evaluation experiments, the bags we use are of size 25-35cm by 40-53cm when laid flat. The color includes red and white with different patterns. (see Figure~\ref{fig-bags}).
The flat workspace has dimensions of 120cm by 180cm.

\subsection{Data collection and processing for training}

In order to train the region perception module, it is necessary to have a dataset that includes labels for the handles and rim. However, labeling these regions in images with crumpled bags is challenging and would require a significant amount of human annotation effort. To address this challenge, we adopt an approach similar to \cite{seita2019deep, qian2020cloth}, where different colors of paint are used to mark the objects.

\begin{figure}[htbp]
	\centerline
	{
		\includegraphics[
		width=0.6\textwidth]{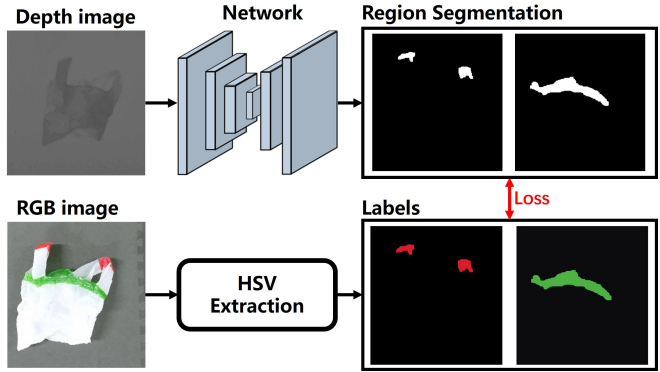}
	}
	
	\caption{\textbf{
		Data collection process and its utilization for training
	}
		Handles and rims are labeled in different colors.
		By color-labeling, the ground truth for the network is provided without the need for expensive human annotations.
		The network receives a depth image as input, while output is the region segmentation results.
		We can get the optimized segmentation network by calculating the loss between the results and the labels
	}
	
	\label{fig-labels}
\end{figure}

In our work, we use the marker pen to mark the handles and rim of the bag with different colors.
It should be noted that the colors cannot be similar to the pattern color of the bag.
Moreover, the marker pen is very friendly to the plastic bag because it can be cleaned by alcohol very easily.
We utilize the red and green colors to mark the bag. Of course, We can use the blue and orange colors to mark the bag, but we only need to modify the corresponding color parameter of the HSV.
Figure~\ref{fig-labels} depicts the data collection process and how applied for training.

We utilize a Microsoft Kinect V2 camera, which can capture RGB-depth information. The camera is positioned above the workspace, providing a top-down perspective.
We collect the data, including RGB pictures and Depth images, from four training bags by taking 2500 images each (resulting in 10,000 total). Our dataset includes bags in various configurations, such as lying on the workspace or being grasped by the robots. We aimed to capture a wide range of volume and orientation configurations to ensure the accuracy and robustness of our perception model.

% \sethlcolor{yellow}
\subsection{Comparison of segmentation network} \label{network-compare}
To optimize the performance of the region perception module, we conduct a comprehensive analysis of various segmentation algorithms. The algorithms compare in this study include U-Net \cite{ronneberger2015u}, PSPNet \cite{zhao2017pyramid}, SegNet \cite{badrinarayanan2017segnet}, and DeeplabV3+ \cite{chen2018encoder}. The U-Net serves as the perception module in one of the baselines, AutoBag \cite{chen2022autobag}, discussed in Section \ref{baselines}. 

We ensure that all four algorithms are configured with the same parameters and trained on the same dataset.
During training, we employ a batch size of 32 and initialize the learning rate to 1e-3. To enhance the diversity of our training data, we apply data augmentation techniques such as random image flipping by 50 percent chance, scaling by 50 percent chance, and rotation by 50 percent chance within the range of [-30 degrees, 30 degrees].

\begin{figure}[htbp]
	\centerline
	{
		\includegraphics[
		width=0.8\textwidth]{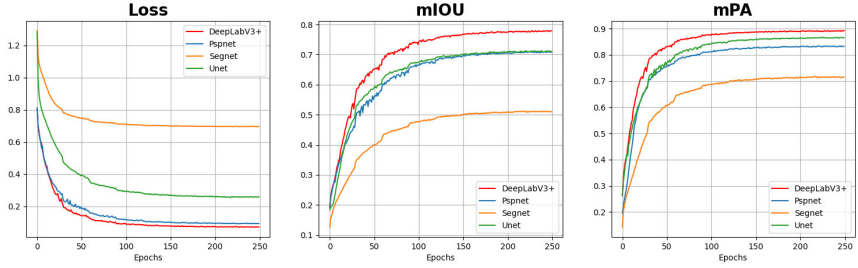}
	}
	\caption{
		Comparing the Learning Curves of Four Region Segmentation Models.
		After 250 epochs, the DeeplabV3+ model achieves remarkable results in our task. It obtains the lowest loss value among all models, with a mean intersection over union (mIOU) of $78.0\%$. Additionally, it demonstrates superior performance in terms of mean Pixel Accuracy (mPA), achieving a score of $89.4\%$, surpassing the other segmentation models in the experiment. These results highlight the performance of DeeplabV3+ compared to the other models
	}
	\label{fig-trainning}
\end{figure}

For the purpose of model training, we adopt an 80-20 train/validation split. All training procedures are conducted on a Ubuntu 18.04 machine equipped with 4 NVIDIA GTX 2080 Ti GPUs, a 3.50 GHz Intel i9-9900X CPU, and 128 GB RAM.

The results depicted in Figure~\ref{fig-trainning} clearly demonstrate the superior performance of DeeplabV3+ compared to the other three segmentation algorithms in our task. 

However, the training time for the four networks varies, as shown in Table~\ref{cost-time}. Among them, SegNet has the shortest training time, followed by U-Net and PSPNet. The limitation of DeeplabV3+ is that it requires the longest training time due to its more complex architecture and a larger number of parameters.

\begin{table}[h!]
	\TBL{\caption{The Training Time of the Four Models\label{cost-time}}}
   {\begin{fntable}\centering
	  \begin{tabular}{@{\extracolsep{\fill}}ccccc}
		\Xhline{1.2pt}
	  \textbf{Metrics}&\textbf{U-Net}& \textbf{PSP-Net}&\textbf{SegNet}&\textbf{DeeplabV3+}\\  
	  \hline
	  \textbf{Time (Hour)} &34&35&27&43\\
	  \Xhline{1.2pt}
	  \end{tabular}
	\end{fntable}}
  \end{table}

Considering these comparisons, we choose to employ the DeeplabV3+ algorithm for our region perception module. This decision is primarily based on the accuracy of detection, which is the main metric we prioritize.  

\begin{figure}[htbp]
	\centerline
	{
		\includegraphics[
		width=0.8\textwidth]{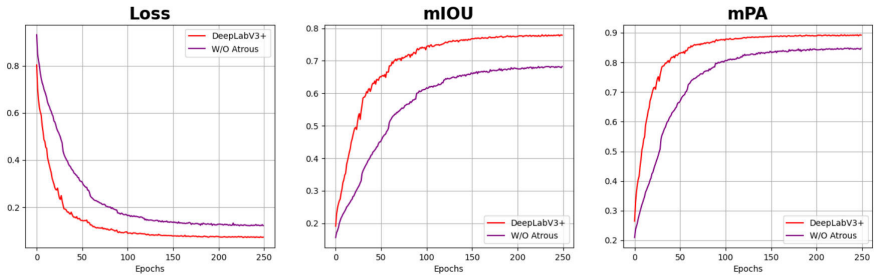}
	}
	\caption{
		The Learning Curves of DeeplabV3+ without Atrous Convolution. After 250 epochs, the loss value of the model without atrous convolution has increased compared to the original DeeplabV3+.The mean intersection over union (mIOU) has decreased from 78.0\% to 68.2\%, while the mean Pixel Accuracy (mPA) decreased from 89.4\% to 83.8\%
	}
	\label{fig-atrous}
\end{figure}

In addition, we analyze the reasons why DeeplabV3+ can achieve the best performance in our task. 
According to the work of Chen et al. \cite{chen2018encoder}, DeepLabV3+ utilizes atrous convolution to merge the coarse features from the lower layers with the fine features from the higher layers, thereby enhancing the segmentation results in terms of edges and details.
To further evaluate the effectiveness of the atrous convolution in our task, we conducted an additional experiment that excluded the atrous convolution, as shown in Figure~\ref{fig-atrous}.

\subsection{Action primitive parameters setting}
In the experiment, we set the parameters of the action primitive. 
Table~\ref{parameters} shows the meaning of each parameter and how they can be obtained. 
It should be noted that the optimal values may vary depending on the particular configuration of the robot and the operating environment.

\begin{table}[htb]
	\TBL{\caption{The Action Primitive Parameters Setting\label{parameters}}}
   {\begin{fntable}\centering
	  \begin{tabular}{@{\extracolsep{\fill}}ccm{5cm}m{5cm}}
		\Xhline{1.2pt}
		\makecell[c]{\textbf{Action}\\\textbf{Primitive}}&\textbf{Parm.}&\textbf{Meaning}&\textbf{Setting}\\  
	\Xhline{1.2pt}
		\multirow{9}{*}{\makecell[c]{\textbf{Bag}\\\textbf{adjustment}}}&$d$& The initial distance between the two grippers after moving to the dangling & Pre-set according to the size of the bags \\
	\cline{2-4}
	&$\Delta d$& The amount of change in the distance between the two grippers each time & Pre-set fixed values according to site configuration \\
	\cline{2-4}
		& $k_s$& The times of horizontal swinging movement & Pre-set fixed values according to site configuration \\
	\cline{2-4}
		&$l$&The horizontal length of swinging movement&Pre-set fixed values according to site configuration \\
	\cline{2-4}
		&$f$&The frequency of swinging movement &Pre-set fixed values according to site configuration \\
	\hline
	\multirow{8}{*}{\makecell[c]{\textbf{Dual-arm}\\\textbf{shaking}}}&$H$&The maximum height that the robot arm can reach during the shaking action&Pre-set according to the capability of the robot. In general, the larger the value, the better\\
	\cline{2-4}
		\multirow{1}{*}&$H^{'}$&The height of grippers after the robot shaking action& Pre-set fixed value and the value should be higher than the bag’s bottom to prevent the bag from touching the surface\\
	\cline{2-4}
		\multirow{1}{*}&$v$&The speed of shaking action&Pre-set according to the max capability of the robot. In general, the larger the value, the better\\
	\hline
		\textbf{\makecell[c]{\textbf{One-arm}\\\textbf{holding}}}& $h$ &The height at which the dual arms descend until the bottom of the bag comes in contact with the workspace&Real-time control based on detecting the lowest point in the middle of the bag through the RGB-D camera\\
	\Xhline{1.2pt}
	  \end{tabular}
	\end{fntable}}
\end{table}

\subsection{Experiment protocol}
\begin{figure}[htbp]
	\centerline
	{
		\includegraphics[
		width=0.6\textwidth]{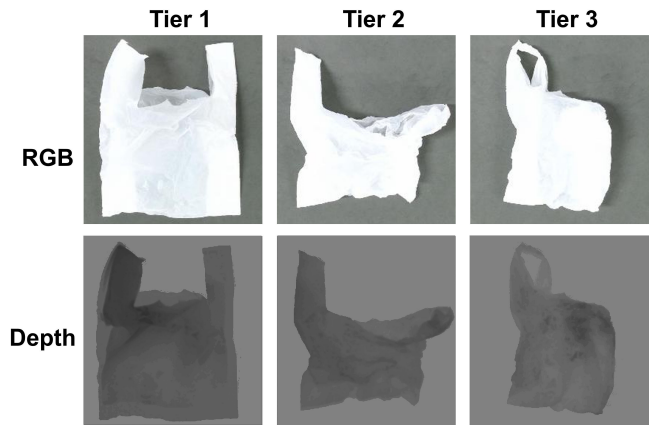}
	}
	\caption{\textbf{Three tiers of initial bag configurations.} The first row shows the RGB images, while the second shows the corresponding depth images
	}
	\label{fig-tier}
\end{figure}

To evaluate ShakingBot, we use four bags, one of which is the middle-size bag from training.  
The other three bags include an unseen-size bag, a different pattern bag, and a pure red color bag.
The goal is to insert 2 identical bottles into each bag(see Figure~\ref{fig-shakingbot}). 
In our definition of a trial, the robot attempts to perform the entire end-to-end procedure: opening a bag, inserting $n$ items into it, and lifting the bag (with items). We allow up to $T=15$ actions before the robot formally lifts the bag. 
The robot will be reset to its home position if it encounters motion planning or kinematic errors during the trial. 
We evaluate the ShakingBot with three difficulty tiers of initial bag configurations (see Figure~\ref{fig-tier}):

\begin{itemize}
	\item Tier 1: The two handles of the bag can be recognized, and the bag has an expanded, slightly wrinkled state lying sideways on the workspace. Besides, the rim has an opening. This requires reorienting the bag upwards.
	\item Tier 2: The tier is similar to tier 1, but the rim area and the degree of wrinkle. The rim area is smaller than tier 1. This will need more actions to enlarge the bag.
	\item Tier 3: There are one or two handles hidden. Another, it has a more complex initial configuration. This requires some actions to expand the bag.
\end{itemize}

At the start of each trial, we perform the initialization process, which involves manipulating the bag so that it falls into one of the tiers.
The trial is considered successful (``Full Succ.''), if the robot lifts the bag off the surface for at least two seconds while containing all items. 
It is a partial success (``Partial Succ.'') if the robot lifts at least one object into the bag during the trial.
In the report, we describe the number of times the robot successfully opens the bag (``Open Bag''), the number of objects that the robot correctly places in the bag opening before lifting (``Placed''), the number of actions which the robot executes (``Actions''), and the time the robot costs (``Time'').

\subsection{Approach comparison} \label{baselines}
We compare ShakingBot to two baselines and two ablations on three tiers.
To evaluate the dynamic shaking action, we compare it against the quasi-static baseline AutoBag \cite{chen2022autobag}, which is the state-of-art in bagging tasks. 
They also used a region perception module, of which architecture is U-Net \cite{ronneberger2015u}, to recognize the key region of the bag. 
While they only used the quasi-static action to manipulate the bag, we adopted the dynamic shaking action.
In the comparison experiment, we utilize AutoBag with Depth images, similar to our training dataset.
To evaluate the region segmentation module, we use an analytical method to detect the handles and rim. 
The analytical method consists of two parts: Harris \cite{harris1988combined} is used to detect the handles, and Canny \cite{canny1986computational} is used to calculate the rim area. Subsequently, we execute the action primitives based on the detection results obtained from the analytical method.
We test these baselines for Tier 1, 2, and 3 configurations. 

\begin{table}[!htbp]
	\TBL{\caption{Results of ShakingBot and Baseline Methods\label{compare}}}
	{\begin{fntable}\centering
			\begin{tabular}{@{\extracolsep{\fill}}cccccccc}
			\Xhline{1.2pt}
		
			\makecell[c]{\textbf{Tiers}}& \textbf{Method}&\makecell[c]{\textbf{Open}\\ \textbf{Bag}}&\textbf{Placed}&\makecell[c]{\textbf{Partial}\\\textbf{Succ.}}&\makecell[c]{\textbf{Full}\\\textbf{Succ.}}& \textbf{Actions}& \textbf{Time (Sec.)}\\  
			\Xhline{1.2pt}

			\multirow{3}{*}{ \textbf{Tier 1}}&{Analytic\&Primitives}&2/8&0.4$\pm$0.7&2/8&1/8&N/A&N/A\\  
			\multicolumn{1}{c}{}&AutoBag&6/8&1.3$\pm$0.9&6/8&3/8&8.2$\pm$3.9									&195.7$\pm$38.8\\  
			\multicolumn{1}{c}{}&ShakingBot&\pmb{7/8}&\pmb{1.6$\pm$0.7}&\pmb{7/8}&\pmb{6/8}&\pmb{7.2$\pm$1.2}	&\pmb{178.5$\pm$15.7}\\  
			\hline

			\multirow{3}{*}{ \textbf{Tier 2}}&{Analytic\&Primitives}&1/8&0.3$\pm$0.7&1/8&1/8&N/A&N/A\\  
			\multicolumn{1}{c}{}&AutoBag&4/8&0.9$\pm$0.9&4/8&3/8&10.8$\pm$2.6									&219.0$\pm$31.8\\  
			\multicolumn{1}{c}{}&ShakingBot&\pmb{6/8}&\pmb{1.5$\pm$0.9}&\pmb{6/8}&\pmb{6/8}&\pmb{8.8$\pm$2.6}	&\pmb{193.6$\pm$29.5}\\  
			\hline

			\multirow{3}{*}{ \textbf{Tier 3}}&{Analytic\&Primitives}&{0/8}& 0.0$\pm$0.0 &0/8&0/8&N/A&N/A\\  
			\multicolumn{1}{c}{}&AutoBag&1/8&0.3$\pm$0.7&1/8&1/8&14.4$\pm$1.5									&258.6$\pm$17.2\\  
			\multicolumn{1}{c}{}&ShakingBot&\pmb{4/8}&\pmb{0.6$\pm$0.8}&\pmb{3/8}&\pmb{2/8}&\pmb{12.5$\pm$5.1}	&\pmb{227.1$\pm$47.9}\\  
			\Xhline{1.2pt}
			\end{tabular}
	\end{fntable}}
\end{table}

We report results on the middle-size bag from the training bag in Table.~\ref{compare}, where we run 8 trials per experiment setting. 
ShakingBot can achieve both a partial success rate and a full success rate on the three respective tiers.
Compared with the traditional analytic method, the analytical way is insufficient for the bagging task. It often fails to detect the right grasping points and fails to calculate the position to place items, especially in complex initial configurations.
Compared with the quasi-static AutoBag, our method can achieve better results with fewer actions and less time, demonstrating dynamic manipulation's efficiency and performance in the bagging tasks.

We additionally analyze the failure modes of ShakingBot. 
Except for the inaccuracy in the prediction and planning errors, there are other failure modes.
For example, when inserting items, the right gripper grasping the items may touch the opened bag, causing the bag to deform, resulting in failure to put the items in the bag.

\subsection{Ablations}

To evaluate the utility of the Bag Adjustment and One-arm Holding primitives, we perform two ablations. 
ShakingBot-A, where the robot does not perform Bag Adjustment action after grasping the two handles.
ShakingBot-H, where the robot does not perform One-Arm Holding to hold the bag, instead releasing one of the handles in the air without the support of the desk surface.
% after enlarging the bag.
We perform these ablations for Tier 2 configuration.

\begin{table}[h!]
	\TBL{\caption{Results of Ablations\label{abltions}}}
	{\begin{fntable}\centering
			\begin{tabular}{@{\extracolsep{\fill}}ccccc}
			\Xhline{1.2pt}
			\textbf{Method}&\makecell[c]{\textbf{Open}\\\textbf{Bag}}&\textbf{Placed}&\makecell[c]{\textbf{Partial}\\\textbf{Succ.}}&\makecell[c]{\textbf{Full}\\\textbf{Succ.}}\\  
			\Xhline{1.2pt}
			ShakingBot-A&2/6&0.8$\pm$1.1&2/6&2/6\\
			ShakingBot-H&0/6& 0.0$\pm$0.0 &0/6&0/6\\
			%		ShakingBot&\pmb{5/6}&\pmb{1.6$\pm$0.7}&\pmb{1.1$\pm$0.8}&\pmb{4/6}&\pmb{3/6}\\
			\Xhline{1.2pt}
			\end{tabular}
	\end{fntable}}
\end{table}

The results are shown in Table~\ref{abltions}. The ablations underperform the full method, demonstrating that Bag Adjustment (Section \ref{bag_adjustment}) as well as One-Arm Holding  (Section \ref{one_arm_holding}) help the bagging tasks.

\subsection{Generalization}

\begin{table}[h!]
	\TBL{\caption{Results of ShakingBot on the Bags with Different Sizes, Patterns, and Colors\label{generalization}}}
	{\begin{fntable}\centering
			\begin{tabular}{@{\extracolsep{\fill}}cccc}
		\Xhline{1.2pt}
		\textbf{Types}&\textbf{Open Bag}& \textbf{Partial Succ.}&\textbf{Full Succ.}\\  
		\Xhline{1.2pt}
		Unseen-size &2/3&2/3&1/3\\
		Unseen-pattern& 1/3 &1/3&1/3\\
		Red Color&2/3& 2/3& 2/3\\
		\Xhline{1.2pt}
			\end{tabular}
	\end{fntable}}
\end{table}

\begin{figure}[!htbp]
	\centerline
	{
		\includegraphics[
		width=0.6\textwidth]{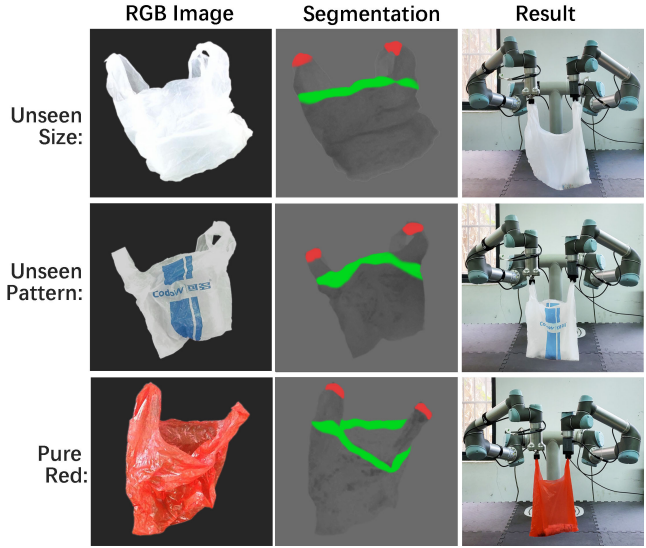}
	}
	\caption{\textbf{Generalization.} Our network is able to generalize to different sizes, patterns, and colors
	}
	\label{fig-generalization}
\end{figure}

In this part, we adopt three bags for the experiments, an unseen size bag, a different pattern bag, and a pure red color bag.
Table~\ref{generalization} presents that ShakingBot can attain 5/9 partial success and 4/9 full success rate.
Our approach can generalize to the bag with different patterns and colors because it only takes depth as input.
It can also perform on the bag of different sizes due to its fully convolutional architecture.
See Figure~\ref{fig-generalization} for visualizations of these generalization experiments.

\section{Conclusion and future work}
In this paper, we propose a novel framework, ShakingBot, for performing bagging tasks, including physical bag opening and item insertion. 
To realize the tasks, we design a novel set of action primitives for dynamic manipulation.
We demonstrate the effectiveness of our method in various bagging tasks, showing its ability to generalize to different sizes, patterns, and colors.
We believe that our method, including our proposed action primitives, can serve as a valuable guide for handling other similar bag-shaped objects or deformable containers.
In the future, we plan to expand our research from plastic bags to encompass other deformable objects.

\begin{con}
	\ctitle{Author Contributions}
	Ningquan Gu and Zhizhong Zhang designed and manufactured the research and contributed equally to this work.
	Ruhan He and Lianqing Yu provided guidance for the research.
	
	\ctitle{Financial Support}
	This study was funded by the project, Research on Video Tracking Based on Manifold Statistical Analysis (No. D20141603), which is the Key Project of the Science and Technology Research Program of the Hubei Provincial Department of Education.

	\ctitle{Conflicts of Interest}
	The authors declare no conflicts of interest exist.
	
	\ctitle{Ethical approval}
	Not applicable.

\end{con}

\bibliographystyle{ieeetr}
\bibliography{reference}
\CTAauthors{N. Gu, Z. Zhang, R. He and L. Yu}
\endctxt{}

\end{document}